\documentclass[sigconf]{acmart}

\renewcommand\footnotetextcopyrightpermission[1]{} 
\usepackage{booktabs}
\usepackage{multirow}
\usepackage{makecell}
\usepackage{appendix}

\begin{document}

\title{Priberam Labs at the NTCIR-15 SHINRA2020-ML: Classification Task}

\author{Rúben Cardoso}
\affiliation{Priberam Labs, Portugal}
\email{ruben.cardoso@priberam.pt}
\author{Afonso Mendes}
\affiliation{Priberam Labs, Portugal}
\email{amm@priberam.pt}
\author{Andre Lamurias}
\affiliation{Priberam Labs, Portugal}
\email{andre.lamurias@priberam.pt}
\newcommand\amm[1]{{\textcolor{blue}{#1}}}

\begin{abstract}
Wikipedia is an online encyclopedia available in $285$ languages. It composes an extremely relevant Knowledge Base (KB), which could be leveraged by automatic systems for several purposes. However, the structure and organisation of such information are not prone to automatic parsing and understanding and it is, therefore, necessary to structure this knowledge. The goal of the current SHINRA2020-ML task is to leverage Wikipedia pages in order to categorise their corresponding entities across $268$ hierarchical categories, belonging to the Extended Named Entity (ENE) ontology.

In this work, we propose three distinct models based on the contextualised embeddings yielded by Multilingual BERT. We explore the performances of a linear layer with and without explicit usage of the ontology's hierarchy, and a Gated Recurrent Units (GRU) layer. We also test several pooling strategies to leverage BERT's embeddings and selection criteria based on the labels' scores. We were able to achieve good performance across a large variety of languages, including those not seen during the fine-tuning process (zero-shot languages).
\end{abstract}

\maketitle
\pagestyle{plain} 

\section*{Team Name}
Priberam Labs - PribL

\section*{Subtasks}

Shinra2020-ML: Classification Task (English, Spanish, French, German, Chinese, Russian, Portuguese, Italian, Arabic, Turkish, Dutch, Polish, Korea, Norwegian, Czech)


\section{Introduction}

Wikipedia is a free online encyclopedia. It is an open-access collection of pages in $285$ languages which are created, edited and maintained by a large community of volunteers, under a system known as open collaboration. The diverse and far-reaching coverage of topics and languages makes Wikipedia a very complete Knowledge Base (KB).

However, Wikipedia was designed and built as a resource for people and it is not trivial to manipulate its information using Artificial Intelligence systems. Therefore, the SHINRA project aims to structure the information contained in Wikipedia in different languages, with the purpose of better leveraging such information with automatic systems.

The SHINRA2020-ML task~\cite{shinra_overview} aims to categorise Wikipedia entities in $30$ languages, based on the Extended Named Entity(ENE) definitions including $200+$ categories~\cite{shinra_overview, ene_website}. These definitions compose a taxonomy with $4$ increasingly specific hierarchical levels which enable a fine-grained typing of entities. For example, "New York" should be classified as "1.5.1.1", where the $4$ layers of hierarchy are Name (1) - Location (1.5) - Geological and Political Entity (1.5.1) - City (1.5.1.1). Note that since the present ontology allows for multi-label classification, an entity can be assigned more than one type.

The training data consists of hand-categorised entities from the Japanese Wikipedia annotated by experts, and language-links associating these Japanese pages to their corresponding Wikipedia page in each one of the $30$ target languages. If the corresponding page does not exist in one of these languages, the page is not considered for that language. A simple step of preprocessing done by SHINRA's organisation leveraged such data to yield, for each language, all the hand-categorised page IDs and their corresponding gold labels.

In this task, we tackle the problem of multilingual multi-label classification. The first challenge is related to the multilingual component of the task: the desired system should be able to achieve good performance in a large set of languages, with variable amount of training data available. This way, the models should be trained with multilingual data and, preferably, should be able to maintain good performance on zero-shot languages. The second challenge arises because a single Wikipedia page can be classified with several categories. Even though $\approx$ 98 \% of the considered pages are assigned with only one category, the classification of the remaining pages poses an additional challenge.

This paper describes our participation at the NTCIR-15 task SHINRA2020-ML. We explore the performance of several systems based on the contextualised embeddings generated by multilingual BERT (mBERT) combined with different pooling strategies and classifiers. The first classifier was a simple linear layer projecting a pooled representation of mBERT's embeddings onto the decision space. This same model was also trained with a small variation in the training data to better leverage the label's hierarchical structure. The second classifier was a Gated Recurrent Units (GRU) layer that sequentially predicted the $4$ hierarchical levels of the ontology. We propose the following contributions:
\begin{itemize}
    \item A study of the performance yielded by different pooling strategies for the embeddings generated by mBERT, using a linear layer as classifier.
    \item A model trained with an extended version of the gold labels which includes the hierarchical parents of leaf categories, with the purpose of better leveraging the ontology's structure.
    \item A model combining mBERT's embeddings with a GRU layer that sequentially predicted the $4$ hierarchical levels of the ontology. Tests were conducted with two different pooling strategies.
    \item These models have proved capable of achieving good performance on multilingual classification across very different languages, including zero-shot languages.
\end{itemize}

\section{Related Work}

The simplest approach to classify texts in multiple languages follows a naive approach which considers the problem as multiple independent problems of monolingual text classification~\cite{evora}. This means that a model is trained for each language with a corpus composed only by texts on that specific language. To classify a given target document, the suitable classifier is selected and then used to predict the appropriate categories. However, this naive strategy fails to take advantage of the corpus' multiliguality and prevents knowledge transfer between different languages, i.e., the model is inherently incapable of leveraging relations learnt from one language and applying them to another.

A possible solution is the use of multilingual word embeddings, capable of mapping words in different languages onto the same vector space and resulting in a language-agnostic model. Extensive literature has been published on this topic~\cite{multiling1, multiling2, multiling3}. An example of such strategy combines multilingual word embeddings with character n-gram features and uses a Support Vector Machine (SVM) as classifier~\cite{svmml}. On this work, monolingual word embeddings are trained for each one of the considered languages and, then, linear mappings between the different monolingual embeddings map them to the same vector space.

More complex classification models can involve convolutional neural networks (CNN), which are able to extract the text's most relevant features and leverage them to predict its categories~\cite{cnn1, cnn2}.  These convolutional layers can also be combined with additional resources, such as auto-associative memory relationships or bidirectional long short-term memory units (LSTMs), and applied to multilingual problems~\cite{cnn_ml1, cnn_ml2}.

Regarding the use of recurrent neural networks (RNN), MultiFit should be highlighted~\cite{multifit}. This model's architecture consists of $4$ quasi-recurrent neural network (QRNN) layers followed by an aggregation layer and $2$ linear layers. First, a Language-Agnostic Sentence Representation (LASER) model~\cite{laser} is trained on the text classification task with multilingual data, while a second QRNN based model is solely pretrained on the target language. Then, the label predictions from the LASER multilingual model are used to fine-tune the pretrained monolingual model on the text classification task. This final model shows very good zero-shot performance, even for very low resource languages. Nevertheless, this approach requires an individual model pretrained on each language, which is computationally expensive and impractical.

Recently, the use of transformer-based language models pretrained on very large corpora has advanced the state of the art in several Natural Language Processing tasks. The most widely used of these models is BERT, Bidirectional Encoder Representations from Transformers~\cite{bert}, capable of yielding contextualised embeddings for the input tokens and an embedding for the whole input sequence. A pooling strategy can then be applied to combine these embeddings and, typically, a final linear layer acts as classifier for the specific downstream task. This strategy results in great flexibility and performance. For text classification, usually, only the embedding representing the whole input sequence is used. Furthermore, the capabilities of such approach have been extended to multilingual tasks by Multilingual BERT (mBERT). This model presents the same architecture as the original BERT but it is trained on a large multilingual corpus containing $104$ languages. The resulting model is capable of achieving impressive performance on multilingual tasks, including under a zero-shot paradigm~\cite{mbert}. 
Performance comparisons between the recent BERT models and the more classical approaches using RNNs, such as LSTMs, show that, for small training corpora and low resource languages, models based on RNNs tend to outperform BERT based models on text classification tasks~\cite{svmvsbert, multifit}. Despite this performance gain, these models are monolingual and, therefore, require specific training for the target language.
In the present work, we propose several models based on Multilingual BERT embeddings. This strategy enables the fine-tuning of a single model with multilingual data, capable of classifying texts not only on languages contained within the training set but also for zero-shot languages.


\section{Models}

\begin{figure}[h]
    \centering
    \includegraphics[width=0.45\textwidth]{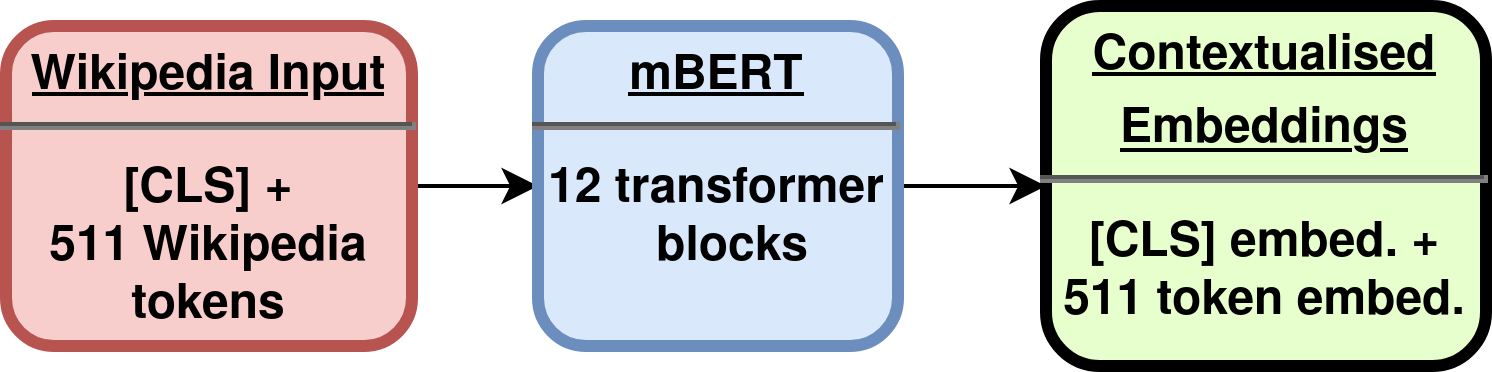}
    \caption{Multilingual BERT is used to obtain the contextualised embeddings corresponding to the first 511 tokens of each Wikipedia page.}
    \label{fig:bert_encode}
\end{figure}

We propose $3$ distinct types of models based on Multilingual BERT's embeddings. For all these models, the first $511$ tokens of each Wikipedia article are concatenated with a preceding $[CLS]$ token, resulting in an input structure for BERT with the form ($[CLS]$, $token_0$, ..., $token_{510}$). BERT yields the contextualised embeddings, with $768$ dimensions, corresponding to each one of the $511$ tokens and the $[CLS]$ token embedding representing the whole input sequence. This encoding structure is represented on figure \ref{fig:bert_encode}. The following proposed models differ in the way they leverage these embeddings and the approach used to handle the ontology's hierarchical structure.

\subsection{Linear Classification}
\label{subsec:lin_classif}

Our first and simplest approach to the current task consists of using a linear layer as classifier. This layer receives a pooled representation of mBERT's output and projects it onto the decision space. The hierarchical structure of the labels was not explicitly leveraged and, therefore, the decision space is composed only by the $193$ leaf labels, i.e., those which correspond to a terminal hierarchical node. To understand the impact of different combinations of token embeddings, three pooling strategies were tested:
\begin{itemize}
    \item mBERT+CLS: only the $[CLS]$ token embedding is used. This embedding passes through a dimension-preserving linear layer ($768$x$768$ dimensions) with hyperbolic tangent activation and a dropout layer with $p=0.1$. The resulting representation, denominated \textit{pooled CLS embedding}, is used as input to the final linear layer (classifier). This approach is represented on figure \ref{fig:bert_cls_pool}. This model is hereafter denominated \textsc{Linear+CLS}.
    \begin{figure}[h]
        \centering
        \includegraphics[width=0.45\textwidth]{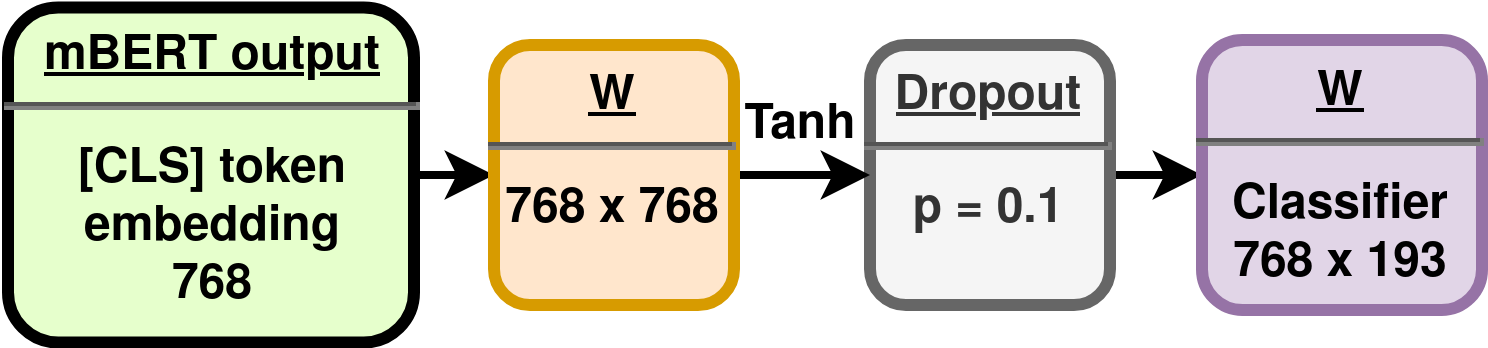}
        \caption{mBERT+CLS pooling strategy and final classifier.}
        \label{fig:bert_cls_pool}
    \end{figure}
    \item mBERT+MEAN: leverages all the $511$ contextualised token embeddings and $[CLS]$ embedding yielded by mBERT through a simple average operation, as shown in figure \ref{fig:bert_mean_pool}. This model is hereafter denominated \textsc{Linear+Mean}.
    \begin{figure}[h]
        \centering
        \includegraphics[width=0.45\textwidth]{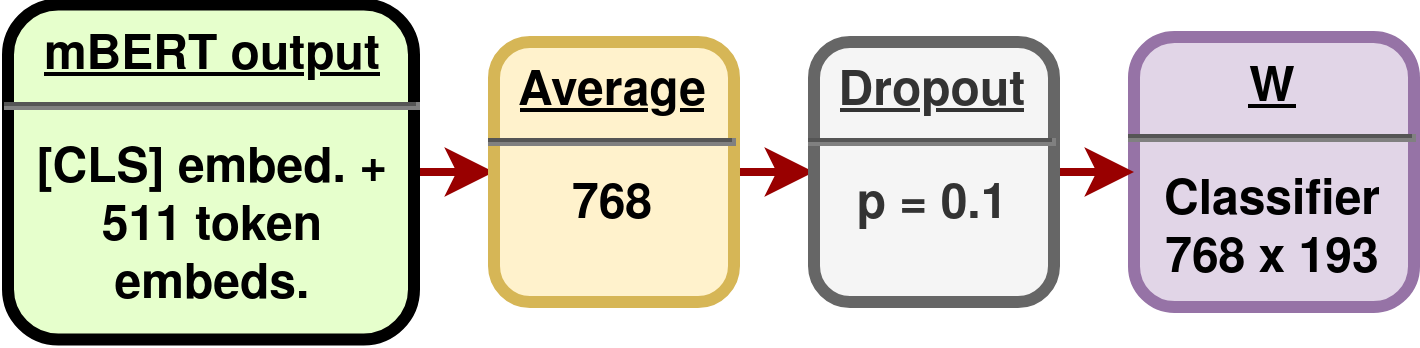}
        \caption{mBERT+MEAN pooling strategy and final classifier.}
        \label{fig:bert_mean_pool}
    \end{figure}
    \item mBERT+CONCAT: this pooling strategy combines concatenation and averaging of different token embeddings. First, the \textit{pooled CLS embedding} is concatenated with the token embeddings corresponding to the first $200$ tokens of each Wikipedia article. This results in a hidden size of $768+200\times768=154368$. Then, the average of the remaining $311$ token embeddings is computed and its result is concatenated to the previous representation, resulting in a final pooled representation with $155136$ dimensions, as represented in figure \ref{fig:bert_concat_pool}. This model is hereafter denominated \textsc{Linear+Concat}.
    \begin{figure}[h]
        \centering
        \includegraphics[width=0.48\textwidth]{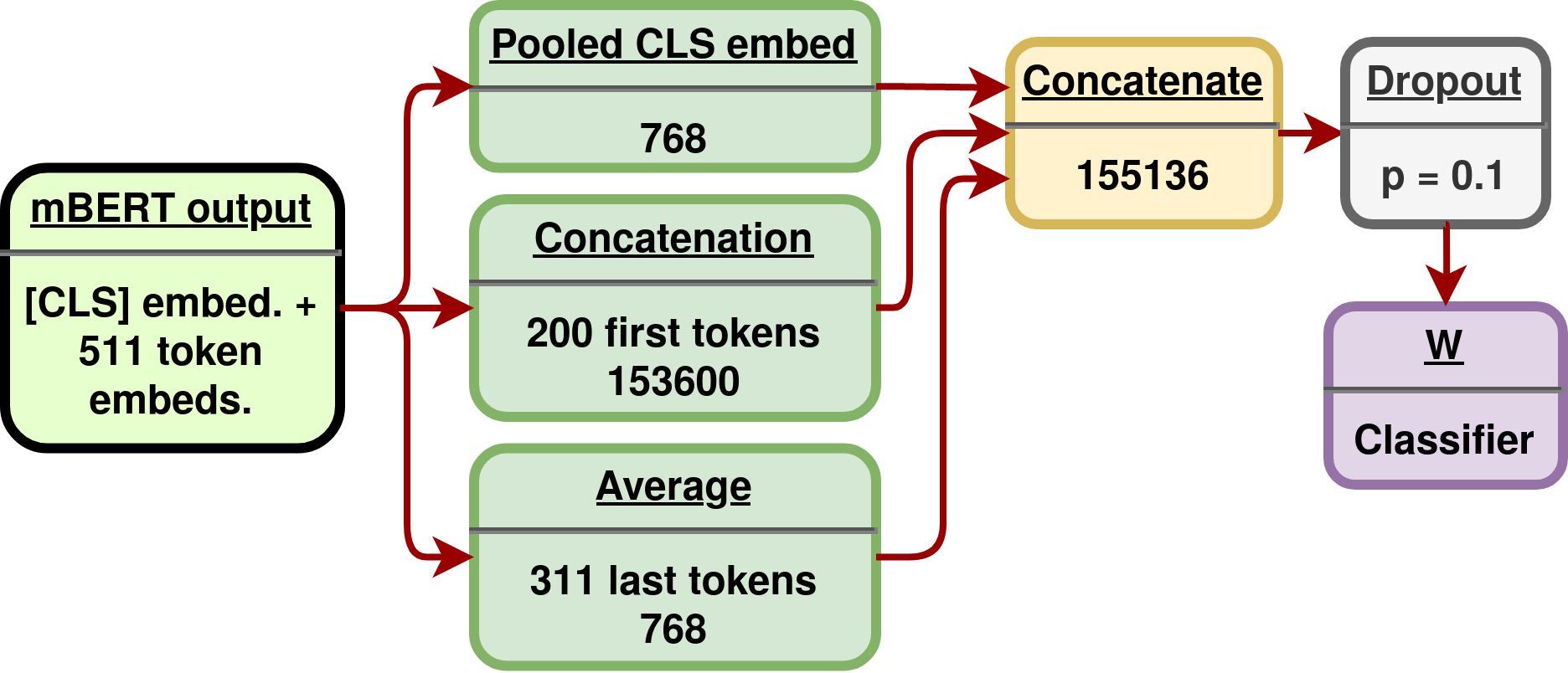}
        \caption{mBERT+CONCAT pooling strategy and final classifier.}
        \label{fig:bert_concat_pool}
    \end{figure}
\end{itemize}

Finally, the linear layer yields a score for each label. To decide whether each one of these labels should be considered, three approaches were tested:
\begin{itemize}
    \item Threshold: a global threshold is fine-tuned on the development set, labels whose scores are above this threshold are considered.
    \item Max Score: given that only about $2$ to $3$ \% of the samples have multiple gold labels, the current task can be approximated as a single-label classification problem without a necessary decrease of performance. This strategy selects for each Wikipedia page the leaf label with maximum score.
    \item Threshold with Max Score: the Threshold approach is applied as before, however, since for some Wikipedia articles all their corresponding label scores may be below the global threshold, this approach leaves some samples without any labels. For these cases, the current strategy performs an additional step which assigns them their maximum scored label.
\end{itemize}

\subsection{Multi-level Hierarchical Classification}

This model presents the same architecture and pooling strategy as the previous \textsc{Linear+Concat} model, represented in figure \ref{fig:bert_concat_pool}. However, the gold labels used during the training process differ from the ones used in the previous classifiers. 

To leverage the hierarchical structure of the Extended Named Entity ontology, the gold labels were decomposed into their hierarchical ancestors and the resulting set of labels became the new gold label set. The problem remains a multi-label classification problem, however, the number of labels per Wikipedia article increases considerably. For example, if one of the labels assigned to a sample is "1.10.4.1" (Fungus), the new set of labels used for training is ["1" (Name), "1.10" (Natural\_Object), "1.10.4" (Living\_Thing), "1.10.4.1" (Fungus)]. This strategy allows the model to learn not only the leaf labels but also the hierarchical steps that lead up to such labels. Note that, with this approach, the decision space includes all the $268$ topology labels.

As for the previous models, during test time a score is yielded by the linear layer and a global score threshold is fine-tuned to decide whether or not a label should be included. However, for this model, an additional step is required: if one of the predicted labels is not a leaf label, its hierarchical descendent with the highest score is selected as part of the predicted label set. This process is repeated until all the labels in the predicted set correspond to leaf labels. The present model is hereafter denominated \textsc{Multi-level Hierarchical}.

\subsection{Hierarchical Sequential Classification}

The present classification approach explicitly leverages the ontology's hierarchical structure. A Gated Recurrent Units (GRU) layer with a hidden size of $768$ dimensions is used to sequentially predict the labels corresponding to the $4$ hierarchical levels. This approach also approximates the task as a single-label classification problem.

In more detail, a loop executes $4$ steps of the GRU layer. Starting from the more general first hierarchical label, at each consecutive step, an additional more fine-grained label is predicted. A masking system enforces the hierarchical structure by reducing the set of possible labels at each step to those corresponding to direct descendants of the label selected in the previous hierarchical level.

\begin{figure}[h]
    \centering
    \includegraphics[width=0.48\textwidth]{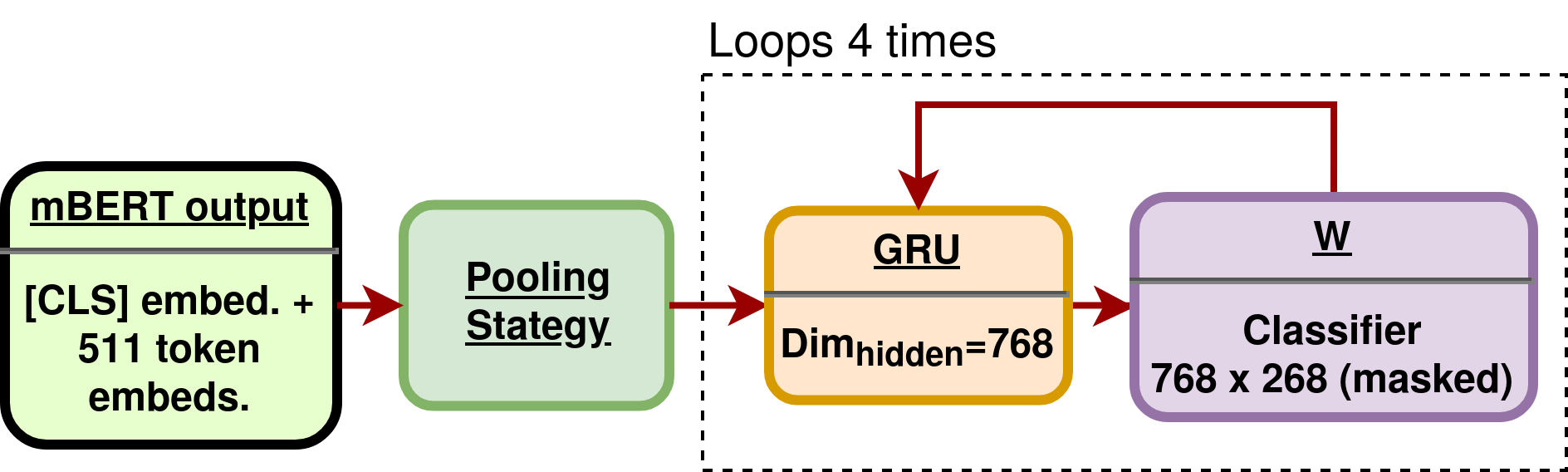}
    \caption{Hierarchical sequential classifier architecture. The GRU's input dimension depends on the pooling strategy used.}
    \label{fig:bert_gru}
\end{figure}

Regarding this model's architecture, it presents a general structure similar to the previous models with the difference that before the final linear layer a Gated Recurrent Units (GRU) layer is added. A scheme of such structure is shown in figure \ref{fig:bert_gru}.

The key part of this model is the GRU/Classifier loop: the GRU's input is the concatenation of the pooling output with the embedding of the label predicted in the previous step. This embedding is the line of the classifier's matrix corresponding to such a label. For the first loop step, a previously predicted label is not available and, therefore, a trainable initial embedding is used. 

Concerning the pooling strategies, two options were tested: mBERT+CLS and mBERT+CONCAT, as described in section \ref{subsec:lin_classif}. These options result in the models hereafter denominated \textsc{GRU+CLS} and \textsc{GRU+Concat}, respectively.

\section{Experimental Setup}

In this section, we describe how the data provided by SHINRA's organisation was used and also report several details regarding the model's implementation and training.

\subsection{Data}
Out of the complete set of $31$ languages with available annotations, we selected the following $13$ as training data for all our models: English (EN), German (DE), Spanish (ES), French (FR), Italian (IT), Portuguese (PT), Russian (RU), Turkish (TR), Arabic (AR), Chinese (ZH), Polish (PL), Dutch (NL), and Korean (KO). This selection takes into account not only the total number of annotated pages per language but also the intention of leveraging the most spoken languages in the world and having a considerable variability of writing systems. During the training process, the articles corresponding to each language were split into $10$ equal slices, each containing $10$\% of the articles in that language. The model is trained with the first slices of all languages, then the second slices, and this process is repeated until all the slices have been used.

The $13$ selected languages contain a total of $21M$ Wikipedia pages, out of which $3.1M$ are annotated. These annotated pages were randomly split into two sets: $95$\% as training set and the remaining $5$\% as a development set used to evaluate the model's performance.
 
An additional leaderboard set with $2000$ samples per language was released. 
This set does not contain the gold annotations, however, it enables the submission of its corresponding predictions on a public leaderboard\footnote{Shinra2020-ML Leaderboard: \url{https://www.nlp.ecei.tohoku.ac.jp/projects/AIP-LB/task/shinra2020-ml}}, which yields scores for the micro precision, recall and F1 metrics. The leaderboard set is not the official test set used for the task, it simply allows public comparison of model performances during the development period.

For some models, we evaluated the performance on this test set in a zero-shot paradigm, i.e., on languages which we did not include in our training set. These zero-shot languages are Norwegian (NO), Danish (DA), Czech (CS), Ukrainian (UK), Vietnamese (VI), and Hindi (HI).
Given that Shinra2020-ML is a shared-task with the purpose of classifying all the Wikipedia articles for several languages, the last relevant set for this task is the complete Wikipedia dump for the languages to be submitted. These languages consist of the $13$ selected training languages and, for one model, also Czech and Norwegian. From this Wikipedia dump set, a subset of annotated articles for each language composes the official test set used to rank the final model's performance.

\subsection{Training and Hyper-parameters}

All models were implemented using the Python packages Transformers and PyTorch~\cite{transformers_package, pytorch}. The contextualised embeddings were obtained from the pretrained model BERT-base multilingual cased.

The training was performed with a maximum sequence length of 512, a batch size of 32, and a maximum learning rate of $2\times 10^{-5}$ following a linear warm-up strategy with $10000$ warm-up steps. The models were trained on a GPU NVIDIA Quadro RTX 8000, with an epoch taking $\approx1$ day.

\section{Results}
It is important to understand the several metrics presented and how the different datasets used for evaluation affect these metrics values. We show scores corresponding to both micro and macro averages. For the micro average, metrics are calculated globally by counting the total true positives, false negatives and false positives. This average is affected by the dataset's distribution of labels, inherently assigning more weight to those with larger frequency. For the macro average, metrics are calculated for each label, and then their unweighted mean is computed. This results in scores which are independent of the set's label distribution.

Table \ref{tab:bert_concat_init} shows the values obtained for different metrics computed for both development and leaderboard sets. These results correspond to the best performances yielded by the \textsc{Linear+Concat} model for the English (EN), Portuguese (PT) and Korean (KO) languages. 

These results show that, independently of the language, the micro scores computed for the leaderboard set are considerably smaller than those corresponding to the development set. Given that this behaviour is verified for all languages and for several samplings of the development set, we can conclude that these two sets have different distributions of labels and, very likely, labels for which the model shows worse performance are much more represented on the leaderboard set. This way, to ensure that our final models perform well for the majority of the labels and, consequently, across datasets with different label distributions, we train the models with the goal of maximising the macro F1 score on the development set.

\begin{table}[H]
\begin{tabular}{@{}c|cccc@{}}
\toprule
Set & Metric & EN & PT & KO \\ \midrule \midrule
\multirow{6}{*}{Dev} & MaF1 & 0.5701 & 0.5355 & 0.5415 \\
 & MaR & 0.5641 & 0.5373 & 0.5486 \\
 & MaP & 0.5836 & 0.5382 & 0.5444 \\
 & uF1 & 0.9586 & 0.9601 & 0.9494 \\
 & uR & 0.9546 & 0.9474 & 0.9440 \\
 & uP & 0.9626 & 0.9732 & 0.9549 \\ \midrule
\multirow{3}{*}{LeadB} & uF1 & 0.733 & 0.696 & 0.720 \\
 & uR & 0.745 & 0.726 & 0.738 \\
 & uP & 0.726 & 0.681 & 0.711 \\ \bottomrule
\end{tabular}
\caption{\textsc{Linear+Concat} model scores corresponding to the number of training steps and score threshold with best performance for the development and leaderboard datasets.}
\label{tab:bert_concat_init}
\end{table}

\begin{figure}[H]
    \centering
    \includegraphics[width=0.48\textwidth]{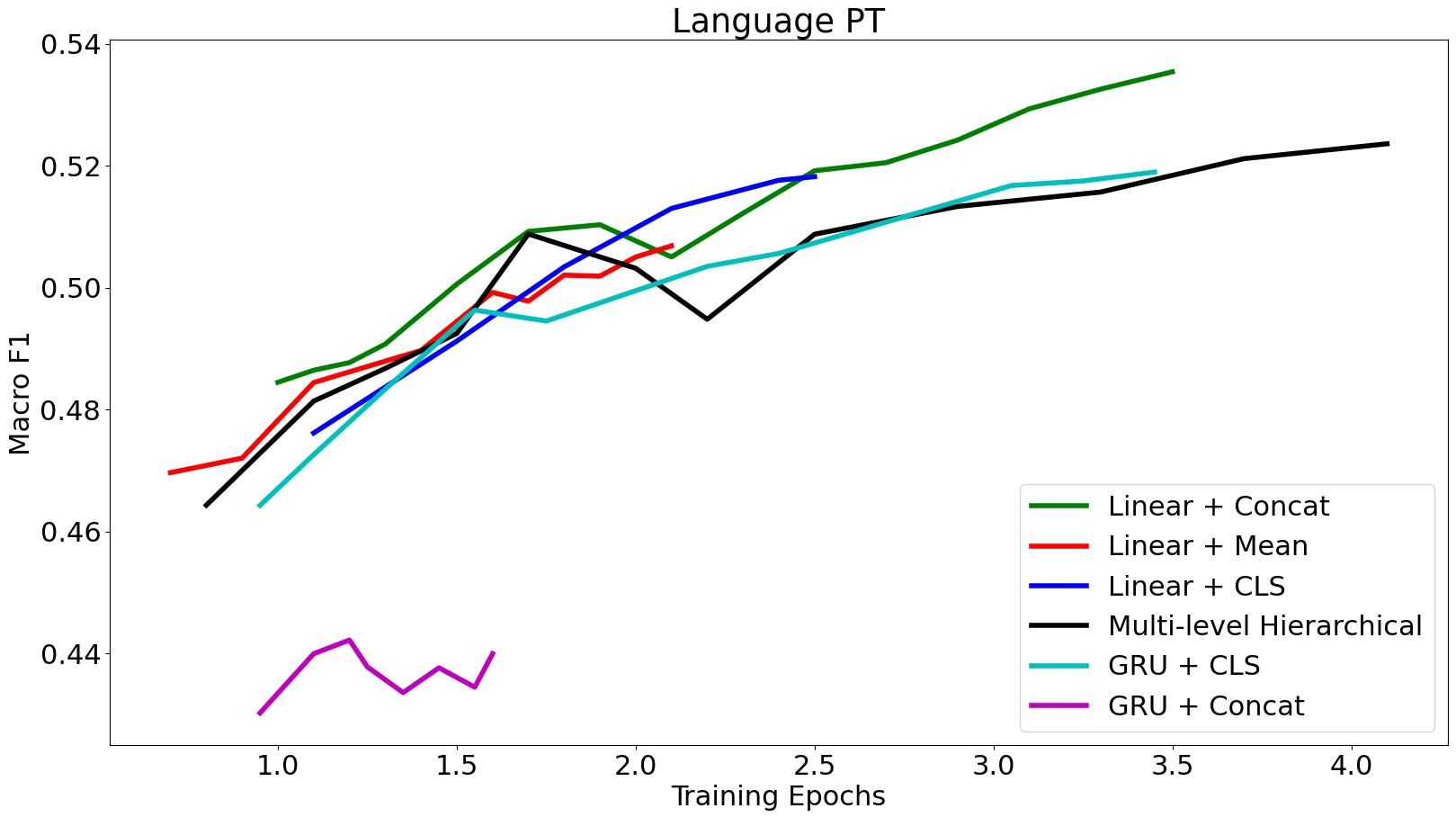}
    \caption{Macro F1 score evolution throughout training for Portuguese.}
    \label{fig:evol_pt}
\end{figure}

\begin{figure}[H]
    \centering
    \includegraphics[width=0.48\textwidth]{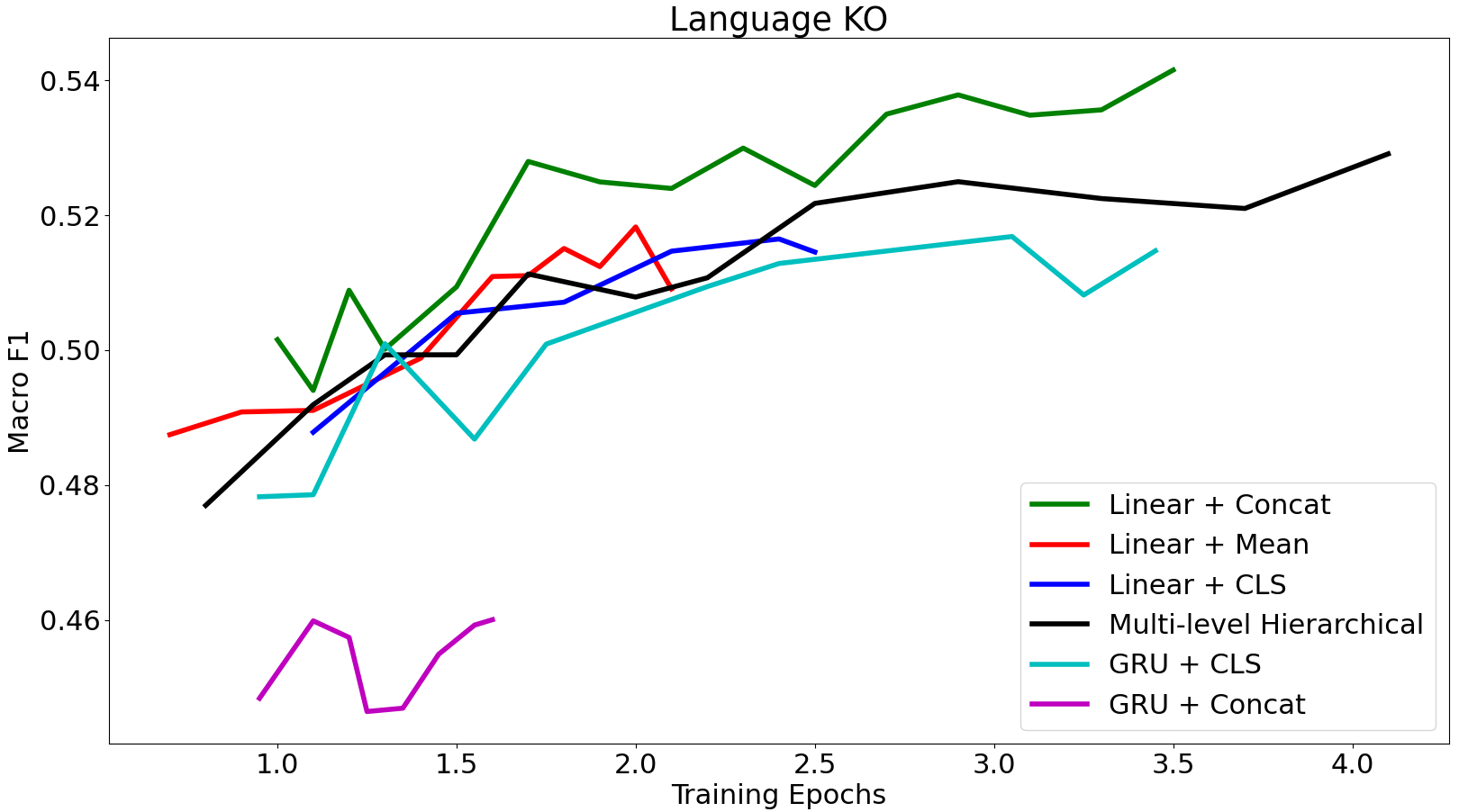}
    \caption{Macro F1 score evolution throughout training for Korean.}
    \label{fig:evol_ko}
\end{figure}

The multilinguality of the current task, arising from the need to classify Wikipedia pages in different languages, implies that the models' performance must be evaluated individually for each language. Different languages can have different best-performing models corresponding to different score thresholds and number of training steps. To properly understand the evolution of the models' performance scores for each language throughout the training process, plots such as those shown in figures \ref{fig:evol_pt} and \ref{fig:evol_ko} were generated for all the $13$ selected training languages. Each model was, in general, trained for $2$ to $4$ epochs and the plots show that quite possibly these could be further trained with additional performance gains. It is also clear that generally the macro F1 scores increase at a similar rate during the training process, maintaining their relative performances. Due to computational limitations, the models were stopped early or were trained for less epochs to allow the training of more promising models.

Table \ref{tab:macrof1_dev} shows the macro F1 scores evaluated on the development set. The values shown correspond to the best-performing number of training steps for each model, and for the Linear and Multi-level Hierarchical models also the best-performing score threshold. Such parameters can be found in table \ref{tab:append_params} under appendix \ref{sec:append}. Table \ref{tab:macrof1_dev} shows that the best pooling strategy when using a linear layer as classifier is \textsc{Concat}. These results were expected given that this model not only leverages the \textit{pooled CLS embedding} but also the individual embeddings of the first, and therefore most relevant, tokens.

The model \textsc{Linear+Concat} is also the best-performing model for most of the languages, only for Chinese and Italian is the \textsc{Multi-level Hierarchical} model capable of very slightly outperforming it.

Regarding the \textsc{Hierarchical Sequential} models, which sequentially predict the $4$ hierarchical label levels using a GRU, it is interesting to note that the \textsc{CLS} pooling can achieve performances very similar to the remaining models while the \textsc{Concat} pooling performs considerably worse. Given that the only difference is the size of the GRU's input (from $1536$ dimensions for \textsc{CLS} to $155904$ for \textsc{Concat}), it is possible that the GRU was not able to leverage such a large input.
\begin{table}[H]
\begin{tabular}{@{}lllllll@{}}
\toprule
 & \thead[l]{\textsc{Linear+} \\ \textsc{Concat}} & \thead[l]{\textsc{Linear+} \\ \textsc{Mean}} & \thead[l]{\textsc{Linear+} \\ \textsc{CLS}} & \thead[l]{\textsc{Multi-level} \\ \textsc{Hierarchical}} & \thead[l]{\textsc{GRU+} \\ \textsc{CLS}} & \thead[l]{\textsc{GRU+} \\ \textsc{Concat}} \\ \midrule \midrule
EN & \textbf{0.5701} & 0.5494 & 0.5528 & 0.5669 & 0.5572 & 0.4790 \\
ES & \textbf{0.5510} & 0.5198 & 0.5271 & 0.5408 & 0.5282 & 0.4557 \\
FR & \textbf{0.5523} & 0.5174 & 0.5176 & 0.5433 & 0.5190 & 0.4597 \\
DE & \textbf{0.5397} & 0.5145 & 0.5185 & 0.5378 & 0.5339 & 0.4616 \\
ZH & 0.5499 & 0.5225 & 0.5317 & \textbf{0.5505} & 0.5360 & 0.4693 \\
RU & \textbf{0.5246} & 0.5061 & 0.5030 & 0.5200 & 0.5029 & 0.4433 \\
PT & \textbf{0.5355} & 0.5069 & 0.5182 & 0.5236 & 0.5190 & 0.4422 \\
IT & 0.5397 & 0.5164 & 0.5238 & \textbf{0.5398} & 0.5262 & 0.4609 \\
AR & \textbf{0.4392} & 0.4320 & 0.4346 & 0.4347 & 0.4153 & 0.3649 \\
TR & \textbf{0.4892} & 0.4756 & 0.4779 & 0.4848 & 0.4734 & 0.4067 \\
NL & \textbf{0.5554} & 0.5226 & 0.5307 & 0.5428 & 0.5347 & 0.4602 \\
PL & \textbf{0.5248} & 0.5048 & 0.5048 & 0.5159 & 0.5106 & 0.4412 \\
KO & \textbf{0.5415} & 0.5183 & 0.5165 & 0.5291 & 0.5168 & 0.4601 \\  \bottomrule
\end{tabular}
\caption{Best macro F1 scores for the development set.}
\label{tab:macrof1_dev}
\end{table}

Table \ref{tab:microf1_leader} shows the micro F1 scores evaluated on the leaderboard set. The models, number of training steps, and score thresholds are the same as those used for table \ref{tab:macrof1_dev}. On this leaderboard set, we have only experimented with the models that achieved best performances on the development set: \textsc{Linear+Concat}, \textsc{Multi-level Hierarchical} and \textsc{GRU+CLS}. Concerning the \textsc{Linear+Concat} model, we explored the performance of the $3$ possible approaches to decide whether or not a label should be considered. Despite their similar performance, the Threshold with Max Score strategy tends to outperform or at least match the other strategies.

On this leaderboard aet, the performances of the \textsc{Linear+Concat} and the \textsc{GRU+CLS} models are more similar, which causes the best-performing model to vary with the considered language. We have additionally evaluated the zero-shot performance of the models that do not require tuning of score thresholds: \textsc{Linear+Concat} Max Score and \textsc{GRU+CLS}. These performances were evaluated for Czech (CS), Ukrainian (UK), Hindi (HI), Vietnamese (VI), Danish (DA), and Norwegian (NO). For both models, the performance slightly decreased for zero-shot languages: average decrease of $3.7$\% for \textsc{Linear+Concat} Max Score and $4.8$\% for \textsc{GRU+CLS}. These zero-shot performances are nonetheless impressive given that these languages present a large variety of writing systems and many of them are considerably different from the languages used during fine-tuning. In general, the \textsc{Linear+Concat} model showed better zero-shot performance than the \textsc{GRU+CLS}.

\begin{table}[H]
\begin{tabular}{@{}llllll@{}}
\toprule
 & \thead[l]{\textsc{Linear+} \\ \textsc{Concat} \\ Threshold} & \thead[l]{\textsc{Linear+} \\ \textsc{Concat} \\ Max score} & \thead[l]{\textsc{Linear+} \\ \textsc{Concat} \\ Threshold \\w/ Max Score} & \thead[l]{\textsc{Multi-level} \\ \textsc{Hierarchical}} & \thead[l]{\textsc{GRU+} \\ \textsc{CLS}} \\ \midrule \midrule
EN & 0.733 & \textbf{0.739} & \textbf{0.739} & 0.713 & 0.707 \\
ES & 0.740 & 0.739 & 0.744 & 0.739 & \textbf{0.751} \\
FR & 0.700 & 0.722 & 0.726 & 0.696 & \textbf{0.735} \\
DE & \textbf{0.758} & 0.754 & \textbf{0.758} & 0.743 & 0.720 \\
ZH & 0.718 & 0.735 & 0.732 & 0.598 & \textbf{0.754} \\
RU & 0.737 & \textbf{0.745} & 0.744 & 0.723 & 0.730 \\
PT & 0.696 & 0.699 & 0.699 & 0.703 & \textbf{0.710} \\
IT & 0.706 & 0.711 & 0.706 & 0.702 & \textbf{0.734} \\
AR & 0.683 & 0.678 & 0.683 & 0.678 & \textbf{0.702} \\
TR & 0.728 & 0.723 & \textbf{0.732} & 0.711 & 0.699 \\
NL & 0.702 & 0.724 & 0.719 & \textbf{0.738} & 0.729 \\
PL & \textbf{0.766} & 0.757 & \textbf{0.766} & 0.701 & 0.722 \\
KO & 0.720 & \textbf{0.746} & 0.731 & 0.721 & 0.738 \\
CS & - & \textbf{0.692} & - & - & \textbf{0.692} \\
UK & - & \textbf{0.696} & - & - & 0.668 \\
HI & - & \textbf{0.605} & - & - & 0.585 \\
VI & - & \textbf{0.722} & - & - & 0.699 \\
DA & - & 0.717 & - & - & \textbf{0.722} \\
NO & - & \textbf{0.717} & - & - & 0.700 \\ \bottomrule
\end{tabular}
\caption{Micro F1 scores for the leaderboard set.}
\label{tab:microf1_leader}
\end{table}

From the results shown in table \ref{tab:microf1_leader}, we selected the two best models to official submit to the Shinra2020-ML task: \textsc{Linear+Concat} Threshold with Max Score and \textsc{GRU+CLS}. We submitted results for our $13$ selected training languages and, for the \textsc{Linear+Concat} model, we have also submitted results for Czech and Norwegian. The micro F1 scores evaluated on the official test set are shown in table \ref{tab:official_test}.

Once again, these two models achieve similar performances for all the languages, with the best model depending on the considered language. Finally, the zero-shot performance of the \textsc{GRU+CLS} model achieves again scores similar to those of the remaining languages.

\begin{table}[H]
\begin{tabular}{@{}lll@{}}
\toprule
 & \textsc{Linear+Concat} & \textsc{GRU+CLS} \\ \midrule \midrule
EN & 0.8012 & \textbf{0.8127 (5th)} \\
ES & \textbf{0.8072 (5th)} & 0.8030 \\
FR & \textbf{0.7852 (3rd)} & 0.7793 \\
DE & 0.7983 & \textbf{0.8024 (5th)} \\
ZH & \textbf{0.7937 (3rd)} & 0.7838 \\
RU & \textbf{0.8308 (2nd)} & 0.8260 \\
PT & 0.8188 & \textbf{0.8236 (2nd)} \\
IT & 0.8189 & \textbf{0.8192 (4th)} \\
AR & 0.7545 & \textbf{0.7627 (1st)} \\
TR & 0.8323 & \textbf{0.8436 (5th)} \\
NL & \textbf{0.8126 (5th)} & 0.8095 \\
PL & \textbf{0.8346 (5th)} & 0.8273 \\
KO & 0.8104 & \textbf{0.8151 (5th)} \\
CS & - & 0.8119 (5th) \\
NO & - & 0.7839 (5th) \\ \bottomrule
\end{tabular}
\caption{Micro F1 scores evaluated on the official test set and corresponding system ranking within the Shinra2020-ML task. The corresponding number of training steps and score threshold can be found in table \ref{tab:append_params} under appendix \ref{sec:append}.}
\label{tab:official_test}
\end{table}

\section{Error Analysis}

To better understand the results obtained and compare the capabilities of the two submitted models, we analysed the mistakes made by each models for Portuguese (PT) and Korean (KO). The results are shown in table \ref{tab:error_types}. We considered the following types of mistakes:
\begin{itemize}
    \item Completely incorrect: zero matches between the predicted and gold label sets.
    \item Over-predicted: predicted set contains at least one correct label, however, additional incorrect labels are also present.
    \item Under-predicted: predicted set contains at least one correct label, however, at least one gold label is missing.
    \end{itemize}
A sample is only considered as correctly predicted if there is a perfect match between predicted and gold label sets.

From table \ref{tab:error_types}, we can notice that the \textsc{Linear+Concat} model shows a considerable number of over-predictions for both languages. On the other hand, as expected, the \textsc{GRU+CLS} model cannot over-predict labels since it only predicts $1$ label per article. However, despite this, the \textsc{GRU+CLS} model shows more incorrect classifications than the \textsc{Linear+Concat} model because the single label chosen by this first model is more often the incorrect one. The number of under-predicted labels is similar across both models and languages.

\begin{table}[H]
\begin{tabular}{@{}lrrrr@{}}
\toprule
\multirow{1}{*}{Model} & \multicolumn{2}{r}{\textsc{Linear+Concat}} & \multicolumn{2}{r}{\textsc{GRU+CLS}} \\ \cmidrule(lr){2-3} \cmidrule(lr){4-5}
 & PT & KO & PT & KO \\ \midrule \midrule
\#correct & 10133 & 8922 & 10025 & 8794 \\
\#incorrect & 546 & 610 & 654 & 738 \\
\#completely incorrect & 384 & 451 & 644 & 727 \\
\#over-predicted & 157 & 155 & 0 & 0 \\
\#under-predicted & 24 & 25 & 21 & 18 \\
\#over and under-predicted & 19 & 21 & 11 & 7 \\ \bottomrule
\end{tabular}
\caption{Error analysis for the two submitted models for Portuguese (PT) and Korean (KO) languages.}\label{tab:error_types}
\end{table}

Table \ref{tab:wrong_labels} shows the labels with smaller $F1$ scores on the development set for the models and languages under analysis. We can see that in general the models struggle to predict the correct fine-grained types of facilities, products and colours. For the Korean language, we additionally notice difficulties related to expressions of time.

\begin{table}[H]
\begin{tabular}{@{}lll@{}}
\toprule
Model & PT & KO \\ \midrule
\textsc{Linear+concat} & \begin{tabular}[c]{@{}l@{}}1.6.1: Facility\_Part\\ 1.7.4: Money\_Form\\ 1.7.10: Offense\\ 1.12.0: Color\_Other\\ 1.12.1: Nature\_Color\end{tabular} & \begin{tabular}[c]{@{}l@{}}1.6.6.0: Line\_Other\\ 1.7.23.0: Title\_Other\\ 1.9.3.0: Natural\_Phenomen.\\ 1.10.5.0: Living\_Thing\_Part\\ 3.8: School\_Age\end{tabular} \\ \midrule
\textsc{GRU+CLS} & \begin{tabular}[c]{@{}l@{}}1.6.3.1: Tomb\\ 1.6.6.4: Water\_Route\\ 1.7.4: Money\_Form\\ 1.7.14: ID\_Number\\ 1.12.1: Nature\_Color\end{tabular} & \begin{tabular}[c]{@{}l@{}}1.6.6.0: Line\_Other\\ 1.7.21.5: Style\\ 1.11.0: Disease\_Other\\ 2.1.1: Time\\ 3.8: School\_Age\end{tabular} \\ \bottomrule
\end{tabular}
\caption{Leaf labels with worse $F1$ performance on the development set.}\label{tab:wrong_labels}
\end{table}

\section{Conclusions}

We can conclude that models based on Multilingual BERT can achieve very good performance across languages with different writing systems and diverse linguistic properties, even under a zero-shot paradigm.

The several tests conducted show that the best pooling strategy for a linear layer classifier involves the concatenation of embeddings corresponding to the first tokens in the text, while for a model with GRU the simple \textit{pooled CLS embedding} results in the best performance.

These two models, \textsc{linear+concat} and \textsc{gru+cls}, yield the best results. Their performances are typically very similar and the best model depends on the considered language. The \textsc{gru+cls} model was additionally capable of maintaining its performance even on zero-shot languages.

\section{Future Work}

Many different tests and experiments have been left for future work due to lack of time. Such future work includes further training all the models until complete score stabilisation, specially those which were stopped very early. Other possible improvements could be the development of a Hierarchical Sequential model capable of multi-label classification, a new input structure and pooling strategy to leverage more than the first $511$ tokens, and experiments with others multilingual language models, such as XLM and XLM-R~\cite{xlm, xlm-r}.

\section{Acknowledgements}

This work is supported by the Lisbon Regional Operational Programme (Lisboa 2020), under the Portugal 2020 Partnership Agreement, through the European Regional Development Fund (ERDF), within project TRAINER (Nº 045347).

\bibliographystyle{ACM-Reference-Format}

\bibliography{shinrabiblio}

\appendix
\section{Submitted Models: Training steps and score thresholds}
\label{sec:append}

\begin{table}[H]
\begin{tabular}{@{}lcrc@{}}
\toprule
\multicolumn{1}{c}{} & \multicolumn{2}{c}{Linear+Concat} & GRU+CLS \\ \cmidrule(lr){2-3} \cmidrule(lr){4-4}
\multicolumn{1}{c}{} & \thead[c]{Training \\ steps} & \thead[c]{Score \\ Threshold} & \thead[c]{Training \\ steps} \\ \midrule \midrule
EN & 321055 & 0.07 & 316503 \\
ES & 284363 & -0.84 & 316503 \\
FR & 302709 & 0.26 & 316503 \\
DE & 266017 & -1.57 & 316503 \\
ZH & 284363 & -0.10 & 279807 \\
RU & 302709 & -0.28 & 316503 \\
PT & 321055 & -0.28 & 316503 \\
IT & 302709 & 0.07 & 316503 \\
AR & 266017 & -0.84 & 316503 \\
TR & 302709 & -0.28 & 316503 \\
NL & 302709 & 0.81 & 298155 \\
PL & 284363 & -0.84 & 316503 \\
KO & 321055 & -0.28 & 279807 \\
CS & 302709 & - & - \\
NO & 302709 & - & - \\ \bottomrule
\end{tabular}
\caption{Number of training steps and score threshold for the submitted \textsc{Linear+Concat} and \textsc{GRU+CLS} models.}\label{tab:append_params}
\end{table}

\end{document}